\documentclass[10pt]{article}

\usepackage[pagewise]{lineno}
\usepackage{multicol}
\usepackage[utf8]{inputenc}
\usepackage{graphicx}
\graphicspath{{figs/}}
\usepackage{subcaption}
\usepackage{amsmath, amssymb}
\numberwithin{equation}{section}
\usepackage{xcolor}
\usepackage{natbib}
\usepackage{bm}
\usepackage{nicefrac}
\usepackage{booktabs}
\usepackage{tabularx,tabulary,array}
\newcolumntype{M}[1]{>{\centering\arraybackslash}m{#1}}
\newcolumntype{N}{@{}m{0pt}@{}}
\usepackage{multirow}
\usepackage{dcolumn}
\usepackage{lscape}
\usepackage[normalem]{ulem}
\usepackage[title]{appendix}
\usepackage{authblk}
\usepackage{hyperref}
\usepackage{adjustbox}
\usepackage{booktabs}

\usepackage{multirow}
\usepackage{rotating,tabularx}

\usepackage[a4paper,top=18mm,right=12mm,left=12mm,bottom=18mm]{geometry}

\usepackage{authblk}
\usepackage[english]{babel}
\usepackage[T1]{fontenc}

\usepackage{amssymb, amsmath, amsthm, blkarray}
\usepackage{graphicx, tikz}

\definecolor{mylightyellow}{rgb}{1,1,.8}
\definecolor{mylightgreen}{rgb}{.8,1,.8}
\definecolor{mydarkred}{RGB}{178,34,34}
\definecolor{mydarkgreen}{RGB}{34,139,34}
\definecolor{mydarkblue}{RGB}{72,61,139}
\definecolor{mydarkyellow}{RGB}{218,165,32}

\usepackage{booktabs}

\usepackage{marginnote}
\usepackage{soul}

\usepackage[]{hyperref}
 \hypersetup{
 colorlinks=true,breaklinks=true,
 urlcolor= mydarkblue,linkcolor= mydarkblue,citecolor= mydarkgreen,
 pdfauthor={S\'aez de Oc\'ariz Borde, H., Sondak, D., Protopapas, P.}
}

\theoremstyle{plain}
\theoremstyle{definition}

\usepackage{eurosym}

\title{Latent Space based Memory Replay for Continual Learning in Artificial Neural Networks}

\author{Haitz S\'aez de Oc\'ariz Borde}
\affil{\small Department of Engineering, University of Cambridge}

\begin{document}

\maketitle

\vspace{5cm}
\begin{abstract}

Memory replay may be key to learning in biological brains, which manage to learn new tasks continually without catastrophically interfering with previous knowledge. On the other hand, artificial neural networks suffer from catastrophic forgetting and tend to only perform well on tasks that they were recently trained on. In this work we explore the application of latent space based memory replay for classification using artificial neural networks. We are able to preserve good performance in previous tasks by storing only a small percentage of the original data in a compressed latent space version.
\end{abstract}

\vspace{2cm}
\noindent {\bf Keywords:} continual learning, memory replay, artificial neural networks, classification.




\twocolumn
\section{Introduction}
\label{sec:introduction}
\setlength{\parskip}{0pt}

Catastrophic forgetting is a recurrent problem in artificial neural networks, and a great handicap for continual learning. Neural networks tend to abruptly forget what they previously learned upon learning new information. The performance of the neural network on a specific task is evaluated based on a loss function and the model updates its parameters to try to optimize for the given loss function without considering its performance in any other task. The location in weight space where the network can reliably perform in a specific task is likely to be distinct from that where it achieved success in a different task. This effectively means that when we retrain a neural network on a different task it will update its weights aiming solely at optimizing the new loss function without minding preserving its performance on other previous tasks.

On the other hand, we humans clearly learn sequentially and build upon previous knowledge to learn new skills. We first learn to walk, then to run, later how to ride a bike... But we do not forget how to walk when we learn to ride a bike. Although it is true that we tend to slowly forget learned information throughout our lifetime (especially if we do not revisit it), new learning does not catastrophically interfere with consolidated knowledge. There are a wide variety of neurophysiological principles that control the interplay between stability and plasticity of the different human brain regions and that help in the development and specialization of the brain based on our sensorimotor experiences throughout life~(\cite{lewk,Murray2016MultisensoryPA,power,zenke}).

In this paper we will focus on memory replay and on how we can draw inspiration from this biological phenomenon to tackle catastrophic forgetting in artificial neural networks. Neural replay sequences were first discovered by studying the hippocampus in rats, but hippocampal replay has also been observed in cats, rabbits, songbirds and monkeys~(\cite{buhry,nokia,Dave2000SongRD,skaggs}). Memory replay has been found to be an important mechanism for learning implemented by biological brains. It consists in firing or replaying neural activity during sleep or wakeful rest so as to reinforce cell activations that occurred beforehand during learning. An entire replay sequence can last a fraction of a second, nevertheless it can play through multiple seconds worth of real life experience. Disrupting brain activity during replay in rodents has experimentally been proven to hinder learning which supports the importance of memory replay~(\cite{EgoStengel2010DisruptionOR,jadhav}).

\section{Memory Replay in Artificial Neural Networks}
\label{sec:Memory Replay in Artificial Neural Networks}

Limited network capacity is not behind catastrophic forgetting in artificial neural networks. The problem resides in how to train the model on different tasks sequentially. Bio-inspired memory replay has been proposed as a tool for continual learning in neural networks by several researchers. We could potentially try to store all previously encountered examples and retrain on them simultaneously when learning a new task, but the scalability of this method is not manageable. Storing previously seen data and alternating it with new data~(\cite{Rebuffi2017iCaRLIC,chaudhry2019tiny}) can be problematic due to both restricted storage capacity and privacy issues. Hence, some researchers have explored using learned generative neural network models of past observations to generate data for replay~(\cite{Robins1995CatastrophicFR,vandeVen2018GenerativeRW,vandeven2020}). This type of approach has also been criticized: for shifting the catastrophic forgetting problem to the training of the generative model~(\cite{schwarz2018progress}). 

Human memory is not perfect, but we surely store memories. Although much is still unknown about how memories are created and accessed within the brain, it is reasonable to suggest that having memories is an evolutionary mechanism to help us survive by making better informed decisions based on past experiences. Having perfect memory is not efficient from a storage perspective and it is not necessary either. We can hypothesize that to make future decisions we only need a compressed version of what we experienced in the past. That compressed version of the experience is a memory that encapsulates the key information of the event without having to store the whole raw input signal. What is more, many times memories relate together information coming from different sensory signals: the visual appearance of a place may remind us of a smell or a song, for example. This may be an indication that we are storing multimodal data in a compressed latent space.

In this paper we will draw inspiration from these suggestions and store previously seen data in latent space which will be used for memory replay. We will first train two neural networks on a given task. The first neural network, which we will call the compressor, will encode the input data into latent space which will be pasted on to a classifier. Initially both networks will be trained and updated simultaneously on a given task (effectively working as a single model). Once we have trained on the first task we will store a fraction of the original data in latent space. Next we will proceed to learn a second similar classification task, but we will freeze the first network. The classifier will be updated based on both the latent space memory and new raw input data that will be pasted through the first (frozen) neural network for encoding. Using this approach we are able to preserve almost the original performance on the first task while learning a second task in a memory efficient way. Latent replay has also been explored in real-time continual learning scenarios~(\cite{latent_replay}). 

The rest of the paper has the following structure. Section~$\ref{Datasets}$ briefly reviews the three datasets used. Section~$\ref{Neural Network Models}$ covers the two neural network models used: the encoder and the classifier. Section~$\ref{Training procedure}$ discusses the training procedure in more detail. In Section~$\ref{Results}$ we show the obtained results and lastly, in Section~$\ref{sec:Conclusion}$ we summarize the final conclusions. 

\section{Datasets}
\label{Datasets}

In this work we use three datasets: the MNIST (\cite{726791,deng2012mnist}), the Fashion-MNIST (\cite{xiao2017fashionmnist}), and the Kuzushiji-MNIST dataset (\cite{Clanuwat2018DeepLF}).

\begin{figure}[!htb]
\includegraphics[width=\linewidth]{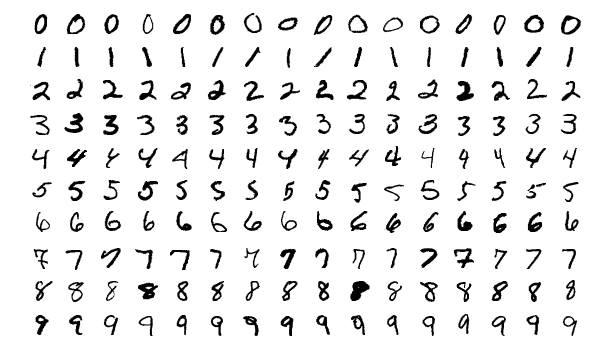}
\caption{Sample MNIST images.}
\label{fig:MNIST}
\end{figure}

\begin{figure}[!htb]
\includegraphics[width=\linewidth]{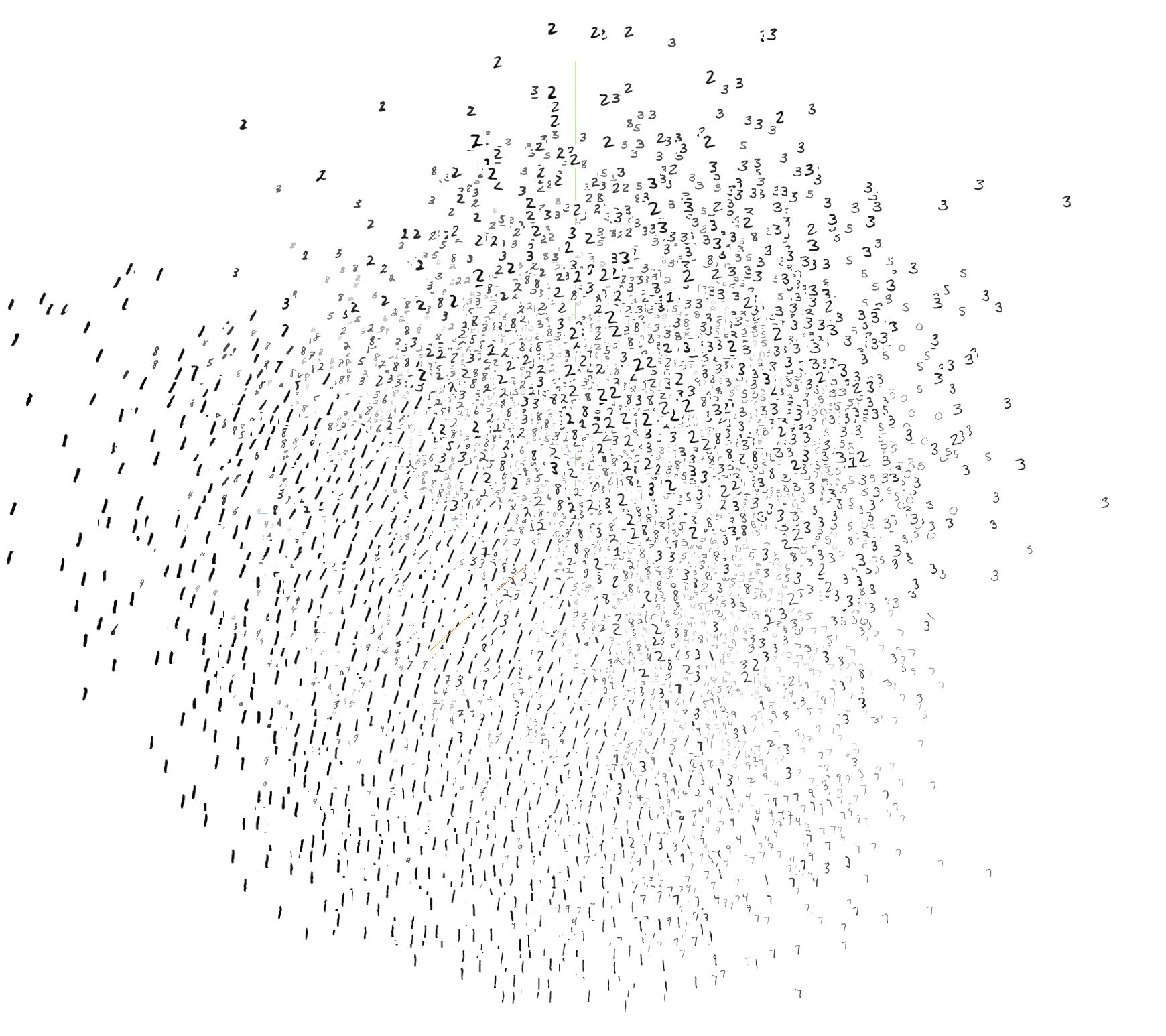}
\caption{MNIST dataset after applying principal component analysis.}
\label{fig:MNIST_pca}
\end{figure}

MNIST is a famous database of handwritten digits, 0 to 9, that is commonly used for training various machine learning algorithms. It was created by remixing samples from the original NIST database~(\cite{Grother1995NISTSD}). MNIST contains a total of 70,000 images, the images are fit into a 28 × 28 pixel bounding box, and have grayscale levels. Figure~$\ref{fig:MNIST}$ and Figure~$\ref{fig:MNIST_pca}$ displays some samples from the dataset. 

\begin{figure}[!htb]
\includegraphics[width=\linewidth]{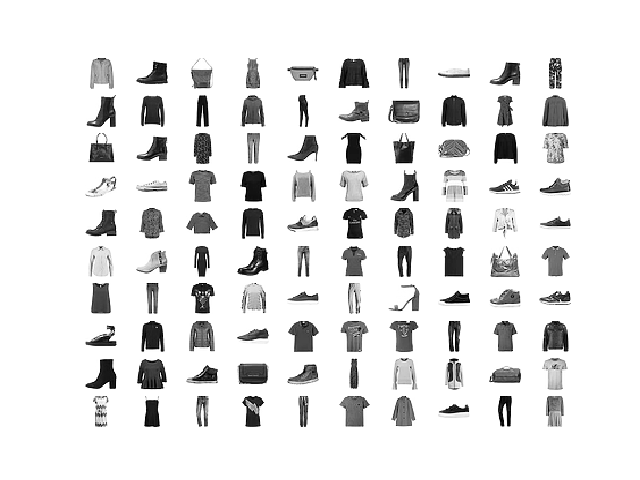}
\caption{Sample Fashion-MNIST images.}
\label{fig:FashionMNIST}
\end{figure}

Fashion-MNIST also comprises 28 × 28 grayscale images of a total of 70,000 fashion products from 10 categories, with 7,000 images per category. It was based on the assortment on Zalando’s website, see Figure~$\ref{fig:FashionMNIST}$. The Fashion-MNIST was intended to be a dropin replacement of MNIST and wanted to provide a more challenging alternative for benchmarking machine learning models.

\begin{figure}[!htb]
\centering
\includegraphics[width=0.8\linewidth]{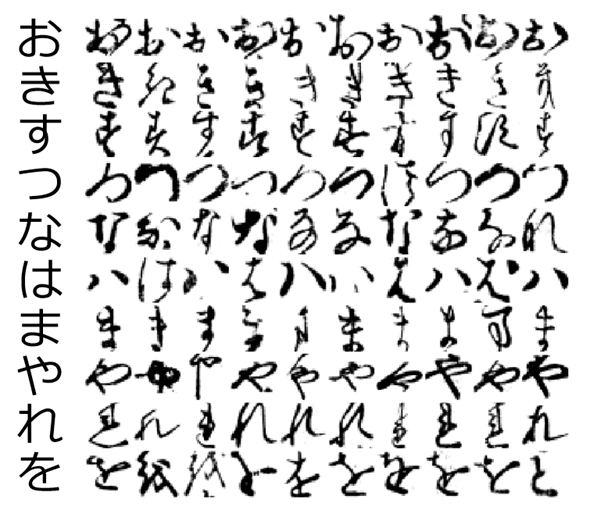}
\caption{Sample Kuzushiji-MNIST images. The 10 classes of Kuzushiji-MNIST are displayed, with the first column showing each character's modern hiragana counterpart.}
\label{fig:kuzuMNIST}
\end{figure}

Lastly, we also work with the Kuzushiji-MNIST dataset which focuses
on Kuzushiji (cursive Japanese). The dataset was created by the National Institute of Japanese Literature as part of a national project to digitize about 300,000 old Japanese books. It contains 70,000 images in 28 × 28 pixel grayscale with 7,000 images per category as well. It has 10 classes with one character to represent each of the 10 rows of Hiragana, see Figure~$\ref{fig:kuzuMNIST}$. Hiragana is a Japanese phonetic lettering system, one component of the Japanese writing system, alongside katakana, kanji and sometimes Latin script. By comparing Figure~$\ref{fig:MNIST}$ and Figure~$\ref{fig:kuzuMNIST}$, it is immediately obvious that the Kuzushiji-MNIST dataset is a more challenging dataset than the original MNIST.

\section{Neural Network Models}
\label{Neural Network Models}

As previously mentioned, we use two neural network models. The first model is a fully convolutional network that maps the input images into the latent space. The network described in Table~$\ref{tab:network1summary}$ maps the input images of dimensions 28 × 28 pixels into a compressed version with 6 × 6 pixels (note that images are in grayscale and both input and output of the network have a single channel). 

The second network is a classifier which has convolutional layers, a fully connected region, and a final log softmax activation function. The architecture is summarized in Table~$\ref{tab:network2summary}$. This second network classifies the output of the first network in 10 classes, that is, it learns to classify the compressed latent version of the original inputs.

Table~$\ref{tab:network1summary}$ and Table~$\ref{tab:network2summary}$ provide detailed information about the number of layers, convolutional channels, the number of neurons used in dense layers, kernel sizes and more. 

\begin{table}[htb!]
\centering
\caption{Description of first neural network: the compressor. Layer 1 has 10 convolutional output channels and Layer 2, 1 convolutional output channel, both use a kernel size of 5. The maxpooling layer uses a kernel size of 3.}
\label{tab:training-prediction cases}
\begin{tabular}{@{}cl@{}}
\toprule
Layer number  & Description           \\ \midrule
1    & Convolutional (2d)    \\
    & Batchnormalization   \\
    & ELU activation  \\
2    &  Convolutional (2d) \\
 & Batchnormalization   \\
  & ELU activation  \\
  3    &  Maxpool (2d) \\
  
\bottomrule 
\end{tabular}
\label{tab:network1summary}
\end{table}

\begin{table}[htb!]
\centering
\caption{Description of second neural network: the classifier. All convolutional layers have 50 convolutional channels and a kernel size of 3. Dense Layer 7 has 16 neurons, and Layer 8 has 10, one for each class.}
\begin{tabular}{@{}cl@{}}
\toprule
Layer number  & Description           \\ \midrule
1    & Convolutional (2d)    \\
    & Batchnormalization   \\
    & ELU activation  \\
2    &  Convolutional (2d) \\
 & Batchnormalization   \\
  & ELU activation  \\
3    &  Convolutional (2d) \\
 & Batchnormalization   \\
  & ELU activation  \\
  4    &  Convolutional (2d) \\
 & Batchnormalization   \\
  & ELU activation  \\
5    &  Convolutional (2d) \\
 & Batchnormalization   \\
  & ELU activation  \\
6    &  Convolutional (2d) \\
 & Batchnormalization   \\
  & ELU activation  \\
  & Flatten\\
7    &  Dense \\
 & Batchnormalization   \\
  & ELU activation  \\
8    &  Dense \\
 & Log Softmax   \\
\bottomrule 
\end{tabular}
\label{tab:network2summary}
\end{table}

\section{Training procedure}
\label{Training procedure}

All datasets used in this work have 10 classes. We will train the classifier on 5 classes first and on the other 5 classes later. For that we organize the data in each dataset in 10 folds, one for each class.

Initially, we train the first and second neural networks at the same time on predicting the first 5 classes. Effectively, at this stage both neural networks work as a single model: the input images go through the first neural network and the outputs are given to the classifier; the weights of both models are updated. We randomly divide the data in the first 5 folds and use $90\%$ for training and $10\%$ for testing.

Once we are done training we freeze the first neural network, which has learned how to "effectively compress" the input data based on the first 5 classes. We pass a small percentage of the original data through the compressor and store the compressed version of the data which will be our latent space based memory. Note that we need to store both the compressed latent space form of the original data and the corresponding labels.

Next we proceed to train the classifier on the remaining 5 classes. The new data is passed through the (frozen) compressor before feeding it to the classifier, whose weights are updated to optimize for the new data and the stored latent space based memory simultaneously. This way we are able to preserve almost the original model performance on the test set used for the first 5 classes while training on the new data.

For training we use a stochastic gradient descent optimizer, a batch size of 1000, a learning rate of 0.001, and the cross entropy loss function. Also, we replay the latent space based memory every time the model is updated on the new data, that is, for every batch we update the model parameters based on the new data and then based on the latent space memory.

\section{Results}
\label{Results}

In this section we explore how much latent space based memory is required to preserve good performance on the original task. In latent space the input images of dimensions 28 × 28 = 784 pixels are represented by just 6 × 6 = 36 pixels ($4.59\%$ the original number of pixels). So, for example, if we were to store $5\%$ of the original data in the latent space based memory that would amount to storing $0.23\%$ of the amount of pixels in original training set. Table~$\ref{tab:training_cases}$ summarizes the different training cases explored in this work.

Table~$\ref{tab:training1performance_mnist}$, Table~$\ref{tab:training1performance_fashionmnist}$, and Table~$\ref{tab:training1performance_kuzushijimnist}$ report the results obtained by the model on the training and testing set when trained on the initial 5 classes for each dataset respectively. 

Table~$\ref{tab:training2performance_mnist}$, Table~$\ref{tab:training2performance_fashionmnist}$, and Table~$\ref{tab:training2performance_kuzushijimnist}$ show the performance of the model on the new classes and also the performance on the original testing data after retraining using memory replay. In the leftmost column of the tables, the training case is specified: the first number refers to the classes used for training and testing (see Table~$\ref{tab:training_cases}$), and the percentage number denotes the amount of latent space memory used. For example, Case $1-0.05\%$ performs the $2^{nd}$ model training based on classes $[5,6,7,8,9]$, stores $1575/31500$ of the original images in latent space, and was originally trained on classes $[0,1,2,3,4]$.

Lastly for comparison, we also include the performance of the model on the original trainining test after retraining on the new classification task when memory replay is not used and the first neural network is not frozen. That is, we retrain the whole model on the new task and check its performance on the original classification classes: catastrophic forgetting occurs. See Table~$\ref{tab:cata_performance_mnist}$, Table~$\ref{tab:cata_performance_fashionmnist}$, and Table~$\ref{tab:cata_performance_kuzushijimnist}$.

\clearpage

\begin{table}[htb!]
\centering
\caption{Training cases explored in this work. More training combinations are possible. The $1^{st}$ training classes column refers to the classes used to originally train the two neural networks (compressor included). The $2^{nd}$ training classes are the classes on which the classifier was additionally trained on using latent space based memory replay. These cases are applicable to all datasets.}
\begin{tabular}{@{}ccc@{}}
\toprule
Case & $1^{st}$ training & $2^{nd}$ training \\ \midrule
 1 & [0,1,2,3,4]  & [5,6,7,8,9] \\
 2 & [5,6,7,8,9]  & [0,1,2,3,4] \\
 3 & [0,1,2,5,6]  & [3,4,7,8,9] \\
 4 & [5,6,7,0,1]  & [8,9,2,3,4] \\
 \bottomrule 
\label{tab:training_cases}
\end{tabular}
\end{table}

Table~$\ref{tab:training1performance_mnist}$, Table~$\ref{tab:training2performance_mnist}$, and Table~$\ref{tab:cata_performance_mnist}$ contain relevant information for the MNIST dataset. By comparing the rightmost columns of Table~$\ref{tab:training2performance_mnist}$ and Table~$\ref{tab:cata_performance_mnist}$ it is clear that using a small proportion of the original data we are able to avoid catastrophic forgetting, whereas without memory replay the model forgets the original classification task.

\begin{table}[htb!]
\centering
\caption{Performance (accuracy) for $1^{st}$ training classes using MNIST dataset.}
\begin{tabular}{@{}ccc@{}}
\toprule
Case & Training & Testing \\ \midrule
 1 &  0.9904 & 0.9885 \\
 2 &  0.9914 & 0.9837 \\
 3 & 0.9901  & 0.9871 \\
 4 &  0.9908 &  0.9881 \\
 \bottomrule 
\label{tab:training1performance_mnist}
\end{tabular}
\end{table}

\begin{table}[htb!]
\centering
\caption{Performance (accuracy) for $2^{nd}$ training classes and performance on the original testing set for MNIST.}
\begin{tabular}{@{}cccc@{}}
\toprule
Case & Training & Testing & Original Testing \\ \midrule
 1 - 0.05 \% & 0.9902  & 0.9685 & 0.9392 \\
 1 - 0.23 \% & 0.9829  & 0.9658 & 0.9555 \\
 1 - 1.15 \% & 0.9897  & 0.9667 & 0.9787 \\
 2 - 0.05 \% & 0.9906  & 0.9891 & 0.9594\\
 2 - 0.23 \% & 0.9902  & 0.9905 & 0.9667 \\
 2 - 1.15 \% & 0.9760 & 0.9717 & 0.9658 \\
 3 - 0.05 \% & 0.9908  & 0.9749 &  0.9788\\
 3 - 0.23 \% & 0.9914  & 0.9764 & 0.9800 \\
 3 - 1.15 \% & 0.9712  & 0.9620 & 0.9811 \\
 4 - 0.05 \% &  0.9886 & 0.9686 & 0.8732 \\
 4 - 0.23 \% & 0.9900  & 0.9669 & 0.9569 \\
 4 - 1.15 \% & 0.9904  & 0.9680 & 0.9779 \\
 \bottomrule 
\label{tab:training2performance_mnist}
\end{tabular}
\end{table}

\begin{table}[htb!]
\centering
\caption{Retraining on new task without memory replay for MNIST.}
\begin{tabular}{@{}cccc@{}}
\toprule
Case & Training & Testing & Original Testing \\ \midrule
 1 &  0.9910 & 0.9804 & 0.0423 \\
 2 &  0.9916 & 0.9902 & 0.0269 \\
 3 & 0.9906  & 0.9824 &  0.2003 \\
 4 &  0.9923 & 0.9833 & 0.0142 \\
 \bottomrule 
\label{tab:cata_performance_mnist}
\end{tabular}
\end{table}

Table~$\ref{tab:training1performance_fashionmnist}$, Table~$\ref{tab:training2performance_fashionmnist}$, and Table~$\ref{tab:cata_performance_fashionmnist}$ summarize the results for the Fashion-MNIST dataset for which we achieve a slightly worse performance on the training set. Nevertheless, the latent space based memory replay still gives satisfactory results and we are able to avoid catastrophic forgetting. Note from Table~$\ref{tab:cata_performance_fashionmnist}$ that forgetting is severe when we simply retrain the whole model, whereas in Table~$\ref{tab:training2performance_fashionmnist}$ we see that the performance on the original training set is really similar to the one obtained before retraining.

\begin{table}[htb!]
\centering
\caption{Performance (accuracy) for $1^{st}$ training classes using Fashion-MNIST.}
\begin{tabular}{@{}ccc@{}}
\toprule
Case & Training & Testing \\ \midrule
 1 & 0.9902 & 0.8897 \\
 2 &  0.9900 & 0.9611 \\
 3 &  0.9643 & 0.8825 \\
 4 &  0.9784 & 0.9174 \\
 \bottomrule 
\label{tab:training1performance_fashionmnist}
\end{tabular}
\end{table}

\begin{table}[htb!]
\centering
\caption{Performance (accuracy) for $2^{nd}$ training classes and performance on the original testing set for Fashion-MNIST.}
\begin{tabular}{@{}cccc@{}}
\toprule
Case & Training & Testing & Original Testing \\ \midrule
 1 - 0.05 \% & 0.9811  & 0.9500 & 0.7959 \\
 1 - 0.23 \% & 0.9907  & 0.9477 & 0.8500 \\
 1 - 1.15 \% & 0.9906  & 0.9468 & 0.8831 \\
 2 - 0.05 \% & 0.9204  & 0.8988 & 0.9211\\
 2 - 0.23 \% & 0.9434  & 0.8912 & 0.9426 \\
 2 - 1.15 \% & 0.9110 & 0.8897 & 0.9540 \\
 3 - 0.05 \% & 0.9696  & 0.9546 & 0.8345 \\
 3 - 0.23 \% & 0.9682 & 0.9531 & 0.8600 \\
 3 - 1.15 \% & 0.9680  & 0.9401 & 0.8779 \\
 4 - 0.05 \% & 0.9127 & 0.8931 & 0.8974 \\
 4 - 0.23 \% & 0.9915  & 0.9008 & 0.8722 \\
 4 - 1.15 \% & 0.9873  & 0.9028& 0.9100 \\
 \bottomrule 
\label{tab:training2performance_fashionmnist}
\end{tabular}
\end{table}

\begin{table}[htb!]
\centering
\caption{Retraining on new task without memory replay for Fashion-MNIST.}
\begin{tabular}{@{}cccc@{}}
\toprule
Case & Training & Testing & Original Testing \\ \midrule
 1 &  0.9860 & 0.9566 & 0.0528 \\
 2 &  0.9772 & 0.9008 & 0.0285\\
 3 &  0.9787 & 0.9614 & 0.0011 \\
 4 &  0.9398  &  0.9159 & 0.0022 \\
 \bottomrule 
\label{tab:cata_performance_fashionmnist}
\end{tabular}
\end{table}

In Table~$\ref{tab:training1performance_kuzushijimnist}$, Table~$\ref{tab:training2performance_kuzushijimnist}$, and Table~$\ref{tab:cata_performance_kuzushijimnist}$ the results for the Kuzushiji-MNIST dataset as displayed. Again, memory replay proves successful at helping the network retain its original performance in classifying the first 5 classes using a small amount of latent space based memory. In this case we see a clear drop in accuracy when we use only $0.05\%$ of the original data as compared to using $0.23\%$. This could be attributed to the fact that in the Kuzushiji-MNIST dataset we can find a quite diverse set of images under the same class (Figure~$\ref{fig:kuzuMNIST}$), hence we require more examples in memory to capture all the diversity in the original data.

\begin{table}[htb!]
\centering
\caption{Performance (accuracy) for $1^{st}$ training classes using Kuzushiji-MNIST.}
\begin{tabular}{@{}ccc@{}}
\toprule
Case & Training & Testing \\ \midrule
 1 &  0.9904 & 0.9683 \\
 2 & 0.9902  & 0.9769 \\
 3 &  0.9904 &  0.9663 \\
 4 & 0.9888  & 0.9737 \\
 \bottomrule 
\label{tab:training1performance_kuzushijimnist}
\end{tabular}
\end{table}

\begin{table}[htb!]
\centering
\caption{Performance (accuracy) for $2^{st}$ training classes and performance on the original testing set for Kuzushiji-MNIST.}
\begin{tabular}{@{}cccc@{}}
\toprule
Case & Training & Testing & Original Testing \\ \midrule
 1 - 0.05 \% & 0.9901  & 0.9697 & 0.7282 \\
 1 - 0.23 \% & 0.9904  & 0.9674 & 0.9108 \\
 1 - 1.15 \% & 0.9886  & 0.9626 & 0.9506 \\
 2 - 0.05 \% & 0.9822 & 0.9631 & 0.7825\\
 2 - 0.23 \% & 0.9903  & 0.9665 & 0.9631 \\
 2 - 1.15 \% & 0.9903 & 0.9668 & 0.9663 \\
 3 - 0.05 \% & 0.9901  & 0.9737 &  0.7067\\
 3 - 0.23 \% & 0.9903  & 0.9703 & 0.9083 \\
 3 - 1.15 \% & 0.9900  & 0.9728 & 0.9091 \\
 4 - 0.05 \% & 0.9907 & 0.9668 & 0.7751 \\
 4 - 0.23 \% & 0.9902  & 0.9645& 0.9223\\
 4 - 1.15 \% & 0.9908 & 0.9637 & 0.9525 \\
 \bottomrule 
\label{tab:training2performance_kuzushijimnist}
\end{tabular}
\end{table}

\begin{table}[htb!]
\centering
\caption{Retraining on new task without memory replay for Kuzushiji-MNIST.}
\begin{tabular}{@{}cccc@{}}
\toprule
Case & Training & Testing & Original Testing \\ \midrule
 1 &  0.9851 &  0.9699 & 0.0122 \\
 2 &  0.9852 & 0.9711 & 0.0229 \\
 3 &  0.9859 & 0.9728 & 0.0120 \\
 4 & 0.9868  & 0.9680 &  0.0117 \\
 \bottomrule 
\label{tab:cata_performance_kuzushijimnist}
\end{tabular}
\end{table}

\section{Conclusion}
\label{sec:Conclusion}

It has been experimentally found that memory replay may be a key element to continual learning in biological brains. The brain uses this technique to reinforce cell activation after learning a new task. In this work, we drew inspiration from these findings and applied latent space based memory replay to classification problems using the MNIST, Fashion-MNIST, and Kuzushiji-MNIST datasets. We found that using a small percentage of the original data and storing it in a compressed latent space form is sufficient for memory replay and to preserve a good performance on previous tasks.

We only explored this technique for similar tasks. In the future it would be interesting to apply this current of thought and similar latent space based memory replay approaches to more diverse problems which are not so similar between them. Efficient memory replay will probably require better encoding mechanisms. Nevertheless, we have shown that memory replay is indeed a good approach for tackling catastrophic forgetting and we have managed to do so storing only a very small percentage of the original data.

\bibliography{biblio.bib}
\bibliographystyle{apalike}

\end{document}